\documentclass[final,5p,times,twocolumn]{elsarticle}

\usepackage{graphicx,times,amsmath,amssymb,mathrsfs,subfigure,stfloats,booktabs,url,multirow,algorithmic,pifont,cleveref}
\usepackage[lined,ruled,commentsnumbered,linesnumbered]{algorithm2e}

\journal{Elsevier}

\begin{document}

\begin{frontmatter}

\title{DenoFlow: Flow Matching for SSVEP Denoising under Real Physiological Artifacts}

\author[1]{Zhentao~He\corref{cor1}}
\author[1]{Ziwei~Wang\corref{cor1}}
\author[1]{Dongrui~Wu\corref{cor2}}
\cortext[cor1]{These authors contributed equally to this work.}
\cortext[cor2]{Emails: m202573788@hust.edu.cn (Zhentao He), vivi@hust.edu.cn (Ziwei Wang), drwu@hust.edu.cn (Dongrui Wu). Dongrui Wu is the corresponding author.}

\affiliation[1]{organization={Hubei Key Laboratory of Brain inspired Intelligent Systems, School of Artificial Intelligence and Automation},
addressline={Huazhong University of Science and Technology},
city={Wuhan},
postcode={430074},
country={China}}

\begin{abstract}
Electroencephalography (EEG)-based brain-computer interfaces (BCIs), particularly steady-state visual evoked potential (SSVEP) systems, are highly vulnerable to noise and artifacts, which severely degrade decoding accuracy. Although recent denoising approaches have shown promise, they are fitted without paired ground truth, can settle on reproducing their input, and are optimized on waveform distance alone, which says nothing about whether the output stays decodable. To address these issues, we propose DenoFlow, which casts SSVEP denoising as transport: instead of learning a direct map from a contaminated trial to a clean one, a field network regresses the velocity of the straight path between them, following the rectified-flow formulation, and denoising integrates that field forward from the observation. The field network is an encoder-decoder that sees the contaminated trial at every layer and the path position at its bottleneck, and a classifier trained alongside it supervises the integrated output. Because the observation itself is both the conditioning input and the starting point of the integration, the model never generates a trial from noise, and training reduces to regression, removing the adversarial min-max game. To obtain paired data on datasets with no ground truth, we injected physiological artifacts of the recorded electromyography (EMG) and electrooculography (EOG) signals under a controlled signal-to-noise target. Experiments on two public SSVEP datasets with five popular SSVEP decoders showed that DenoFlow outperformed seven baseline denoising models on both signal fidelity and downstream decoding accuracy. Code is available at \url{https://github.com/wzwvv/DenoFlow}.
\end{abstract}


\end{frontmatter}

\section{Introduction}
A brain-computer interface (BCI) establishes a direct communication pathway between the brain and an external device, allowing users to interact with the environment without relying on peripheral nerves and muscles \cite{Rosenfeld2017}. Among various non-invasive modalities, electroencephalography (EEG) has been widely adopted in BCIs due to its high temporal resolution, portability, low cost, and ease of deployment \cite{wu2020transfer}. EEG-based BCIs have found application in device control \cite{wang2025mvcnet}, cognitive assessment \cite{zheng2017identifying}, clinical diagnosis \cite{wang2025cst}, and neurorehabilitation \cite{Ienca2018NB}.

Steady-state visual evoked potential (SSVEP) \cite{lin2006cca} is a classical paradigm of EEG-based BCIs. In a typical SSVEP-based BCI, multiple visual targets flash at distinct frequencies, phases, or both. Fixating on a specific target modulates neural activities in the user's visual cortex. The system thus decodes the user's attentional target by matching the frequency and phase characteristics of EEG signals to the flashing visual stimuli.

Practical deployment of EEG-based BCIs remains severely constrained by the inherently low signal-to-noise ratio (SNR) of EEG signals. Artifacts and noise frequently contaminate the EEG signals \cite{tatum2011artifact}, including electrooculographic artifacts from eye blinks and eye movements \cite{wang2015removal,shahbakhti2021vme}, electromyographic artifacts caused by facial muscles, cardiac-related artifacts, baseline drift, and electromagnetic interference \cite{chen2025self}. These sources have to be suppressed before a recording can be decoded reliably.

Traditional EEG denoising techniques operate at the signal level, including band-pass and notch filtering, independent component analysis (ICA), canonical correlation analysis (CCA), principal component analysis (PCA), discrete wavelet transform (DWT), and empirical mode decomposition (EMD) \cite{shi2024removal,wang2015removal,chen2017use,ouyang2022handling,shahbakhti2021vme}. Recently, data-driven EEG denoising models have received increasing attention, including convolutional neural networks (CNNs) \cite{sun2020novel,zhang2021eegdenoisenet,zhang2021novel,nagar2022orthogonal}, recurrent neural networks (RNNs) \cite{zhang2021eegdenoisenet,gao2022eeg}, Transformers \cite{chen2023denosieformer,yin2023gan,tang2025ct,jiang2025wnotnet}, and Mambas \cite{chen2025denoisemamba}. Generative models like generative adversarial network (GAN) \cite{sawangjai2021eeganet,dong2023approach,yin2023gan}, variational autoencoder (VAE) \cite{ghosh2024beta,mahaseni2024eeg}, and diffusion \cite{huang2024eegdfus} have also been introduced to denoise signals by generative reconstruction \cite{wang2026synthetic,wang2026comadit}.

Nevertheless, several challenges remain in EEG denoising:
\begin{enumerate}
	\item Traditional denoising techniques such as ICA and CCA rest on strong assumptions about noise independence and signal stationarity, so their effectiveness diminishes on non-stationary or spectrally overlapping artifacts.
	\item Public SSVEP datasets record the contaminated signal only, so a supervised denoiser has no clean counterpart to be trained against. Models fitted to synthetic perturbations instead transfer poorly to the physiological artifacts encountered in practice.
	\item A model trained to map a contaminated trial directly onto a clean one can settle on reproducing its input. The training loss stays low, but the artifact is never separated from the task-related neural response.
	\item Objectives defined on waveform distance alone weigh every sample equally, so a denoiser can suppress the stimulus response together with the artifact. Nothing in such an objective constrains what the downstream decoder finally receives.
\end{enumerate}

\begin{figure}[htpb] \centering
	\subfigure[]{\includegraphics[width=.75\linewidth]{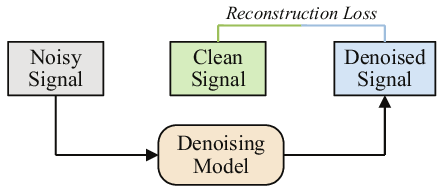}}
	\subfigure[]{\includegraphics[width=.75\linewidth]{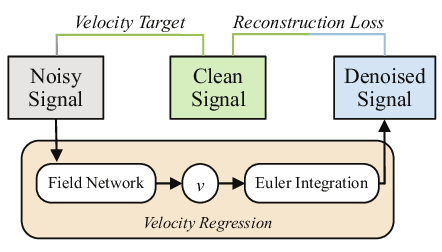}}
	\caption{A traditional EEG denoising model (a) maps a contaminated trial to a clean one in one step, whereas DenoFlow (b) regresses the velocity of the path between them and integrates it forward from the observation.} \label{fig:intro}
\end{figure}

To address the above issues, we propose DenoFlow, which casts SSVEP denoising as learning the path from a contaminated trial to its clean counterpart. A field network is trained to regress the velocity of the straight path between the two, and denoising integrates that field forward from the observation. A classifier trained alongside it supplies a classification loss, so waveform distance is not the only thing shaping the trajectory: The denoised trial also has to remain decodable. The comparison of traditional denoising models and the proposed DenoFlow is illustrated in Figure~\ref{fig:intro}.

The main contributions of this work are summarized as follows:
\begin{enumerate}
	\item We propose DenoFlow, which treats SSVEP denoising as learning the path from a contaminated trial to its clean counterpart rather than a direct mapping between them, and denoises by integrating the learned field forward from the observation. A classifier trained jointly with the field network supervises the integrated output.
	\item We design a field network for this task: The contaminated trial is regarded as a conditioning plane through every layer of an encoder-decoder, while the path position enters once at a Transformer bottleneck, where it describes the trial as a whole rather than any part of it.
	\item We build the paired noised training data by injecting the recorded physiological electromyography (EMG) and electrooculography (EOG) artifacts into clean trials under a controlled signal-to-noise target, which supplies a reference for datasets that provide none.
	\item Experiments on two public SSVEP datasets with five popular SSVEP decoders showed that DenoFlow outperformed seven baseline denoising models on both signal fidelity and downstream decoding accuracy.
\end{enumerate}

The remainder of this paper is organized as follows: Section~\ref{sect:rw} reviews related work. Section~\ref{sect:me} details the proposed DenoFlow. Section~\ref{sect:er} discusses the experimental results and provides analyses. Section~\ref{sect:conclusions} draws conclusions.

\section{Related Work} \label{sect:rw}
\subsection{EEG Denoising}
Traditional approaches focus on denoising at the signal level, including band-pass and notch filtering \cite{shi2024removal} to suppress out-of-band noise and power-line interference; ICA \cite{wang2015removal} to remove EEG artifacts; CCA \cite{chen2017use} to facilitate artifact removal in few-channel EEG;  PCA \cite{ouyang2022handling} to mitigate redundant channel noise; and DWT \cite{shahbakhti2021vme} and EMD \cite{wang2015removal,chen2017use} to isolate noise components. These approaches are limited by strong assumptions and manual feature extraction, encouraging a shift to end-to-end deep learning models.

Data-driven deep learning models have been successfully used in EEG denoising due to their ability to automatically learn discriminative representations. CNNs are widely employed to capture temporal and spatial features of EEG signals. Zhang \textit{et al.} \cite{zhang2021novel} proposed a 1D CNN to ascend feature dimensions and downsample to remove muscle artifacts. Sun \textit{et al.} \cite{sun2020novel} designed a 1D residual CNN for raw waveform-based EEG denoising. Nagar \textit{et al.} \cite{nagar2022orthogonal} proposed a fractional 1D CNN autoencoder to remove muscle artifacts from EEG signals. Zhang \textit{et al.} \cite{zhang2021eegdenoisenet} evaluated the performance of four classic networks, including a fully-connected network, a simple and a complex CNN, and an RNN, on their manually collected EEG dataset. Gao \textit{et al.} \cite{gao2022eeg} proposed a CNN-LSTM model with morphological feature extraction and reinforcement to remove artifacts.

More recently, Transformers and Mambas have also been introduced to EEG denoising, leveraging self-attention and state-space duality mechanisms to capture long-range temporal dependencies of EEG signals. Chen \textit{et al.} \cite{chen2023denosieformer} designed Denosieformer with a residual VAE structure to distinguish contaminated signal patterns. Tang \textit{et al.} \cite{tang2025ct} proposed CT-DCENet, a CNN-Transformer denoising model incorporating a linear ensemble module and an information complementation module. Jiang \textit{et al.} \cite{jiang2025wnotnet} proposed a wavelet-based Transformer architecture, capturing the spectral features and global contexts. Yin \textit{et al.} \cite{yin2023gan} proposed GCTNet, a GAN-guided parallel CNN and transformer network. Chen \textit{et al.} \cite{chen2025denoisemamba} designed a CNN-Mamba denoising model to eliminate artifacts, integrating the CNN with structured state-space duality mechanisms.

In addition to discriminative models, generative models have emerged as powerful tools for EEG denoising, including GANs \cite{goodfellow2014GAN}, autoencoders (AEs), VAEs \cite{kingma2013vae}, and diffusion \cite{ho2020denoising} models. GANs are applied to EEG denoising by enforcing distribution-level similarity between denoised and clean signals \cite{sawangjai2021eeganet,dong2023approach,yin2023gan}. AEs learn compact latent representations to reconstruct clean EEG signals from noisy inputs \cite{gogna2016semi,bethge2022eeg2vec}. VAEs further impose probabilistic constraints on the latent space, enabling more structured feature learning \cite{ghosh2024beta,mahaseni2024eeg}. Huang \textit{et al.} \cite{huang2024eegdfus} proposed a dual-branch conditional diffusion model, utilizing noisy EEG information to guide the clean signal generation. Across these families, the denoiser is fitted without a clean reference to match, can settle on reproducing its input, and is supervised by waveform distance alone, which leaves the decodability of its output unconstrained.

Flow matching offers a different route to generative modeling. Instead of reversing a noising process over many steps as diffusion does \cite{ho2020denoising}, a network is trained to regress the velocity of a path connecting the two distributions, and a sample is obtained by integrating that field \cite{lipman2023flow}. Rectified flow takes the path to be the straight line between its endpoints \cite{liu2023rectified}, so the velocity is constant along it and the integration is accurate within a handful of steps. The formulation has been developed almost entirely for generation, where the path starts from Gaussian noise. Denoising is a more favorable setting for it: the contaminated trial is an observation rather than a sample to be invented, so it can serve as the origin of the path, and the model is never asked to produce a trial from noise.

\subsection{SSVEP Decoding}
Traditional SSVEP decoding algorithms are predominantly based on correlation analysis. Lin \textit{et al.} \cite{lin2006cca} were the first to introduce CCA into SSVEP decoding, which computes the correlation between recorded EEG signals and a set of frequency-specific sine-cosine reference signals. The stimulus frequency corresponding to the maximum correlation is then selected as the target. Bin \textit{et al.} \cite{bin2009online} extended CCA to an online multi-channel SSVEP-based BCI, showing that the method is fast enough to drive a system in real time. Chen \textit{et al.} \cite{chen2015fbcca} introduced filter bank CCA (FBCCA), which applies CCA across multiple sub-bands and combines the results.

Training-based CCA variants are among the most widely used approaches. Pan \textit{et al.} \cite{pan2011enhancing} estimated subject-specific phase information from training data and embedded it into the reference signals. Individual template-based CCA \cite{bin2011high} replaces the analytical references with subject-specific averaged SSVEP templates. Zhang \textit{et al.} \cite{zhang2011multiway} proposed multi-way CCA, which optimizes the reference signals by subject data. Tanaka \textit{et al.} \cite{tanaka2013task} proposed task-related component analysis (TRCA), which maximizes the reproducibility of neural responses across repeated trials at the same stimulus frequency. Nakanishi \textit{et al.} \cite{nakanishi2017trca} subsequently adapted TRCA for SSVEP decoding, effectively suppressing background noise and improving the SNR. Sun \textit{et al.} \cite{sun2021similarity} introduced a similarity constraint into TRCA to better remove noise. Liu \textit{et al.} \cite{liu2021improving} proposed task-discriminant component analysis (TDCA), exploiting informative temporal structures for frequency recognition.

Deep models have become a dominant way for SSVEP decoding due to their powerful ability to learn discriminative features. CNNs, such as EEGNet \cite{lawhern2018eegnet} and complex spectrum CNN (CCNN) \cite{ravi2020CCNN}, extract spatiotemporal representations without relying on handcrafted features. SSVEPNet \cite{pan2022ssvepnet} incorporates long short-term memory (LSTM) layers to model temporal dynamics. To better capture long-range dependencies, hybrid CNN-Transformer models have recently been introduced. SSVEPformer \cite{chen2023ssvepformer} integrates self-attention mechanisms, and DBConformer \cite{wang2025dbconformer} learns temporal and spatial features in two parallel branches. However, SSVEP decoding models remain sensitive to noise and artifacts in EEG signals, and their performance often degrades when EEG quality is compromised, underscoring the necessity of effective denoising strategies.

\section{DenoFlow} \label{sect:me}

\begin{figure*}[htpb] \centering
	\includegraphics[width=\textwidth]{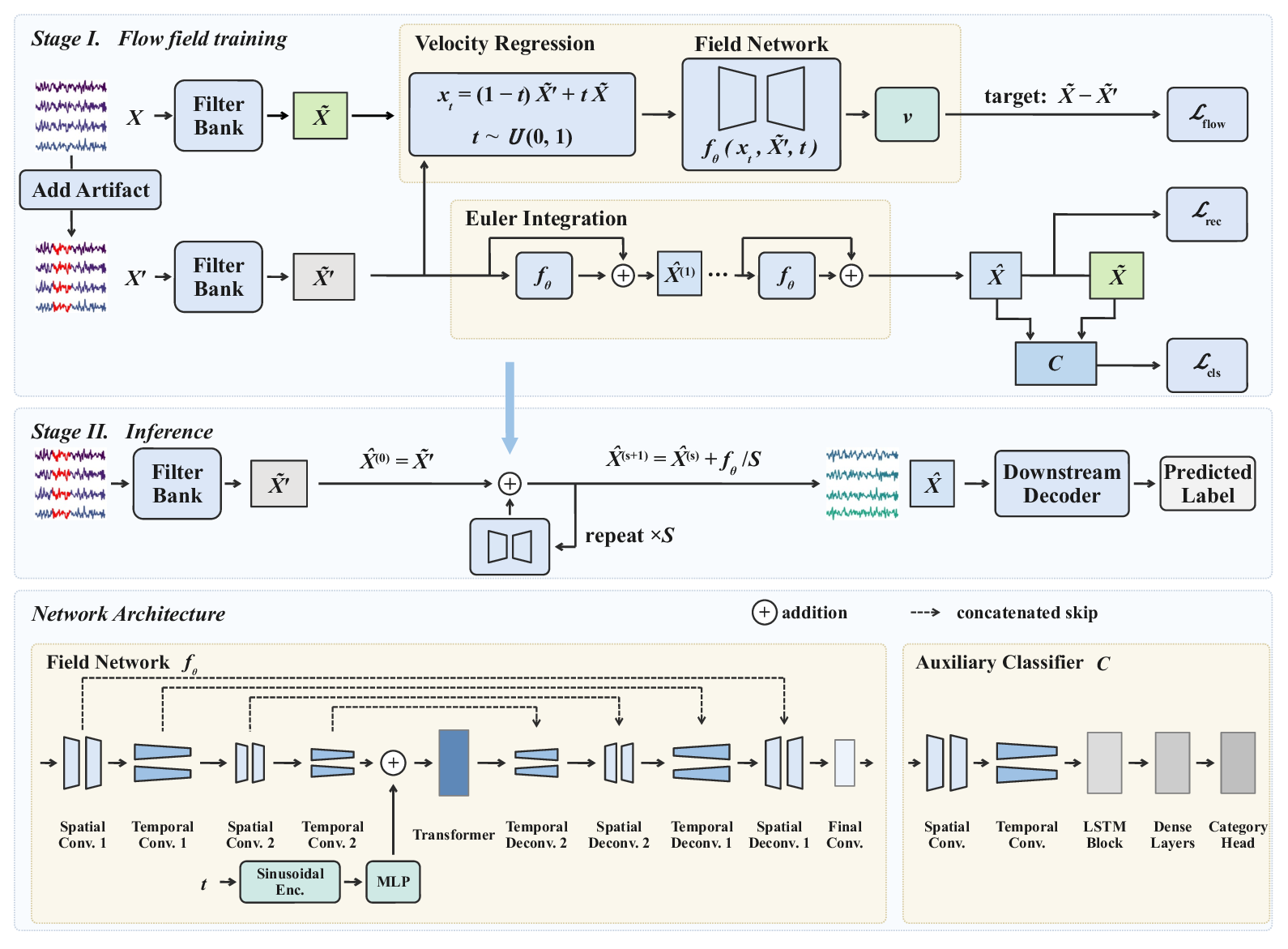}
	\caption{Overall framework of DenoFlow. Stage I trains the field network on injected artifact pairs, with a classifier supervising the integrated output. Stage II integrates the field forward from a contaminated trial.}
	\label{fig:denonet_framework}
\end{figure*}

\subsection{Overview}
Figure~\ref{fig:denonet_framework} illustrates the overall architecture and training pipeline of DenoFlow.

DenoFlow casts denoising as transport rather than as a one-step mapping. The signals involved are the following. An uncontaminated trial $X$ becomes a contaminated trial $X'$ once an artifact is injected into it. Both are then passed through the same filter bank, giving the reference $\tilde{X}$ that the model is trained against and the input $\tilde{X}'$ that it receives. Existing generative denoisers learn a direct map from $\tilde{X}'$ to $\tilde{X}$, trained adversarially so that the outputs match the distribution of the reference \cite{yin2023gan,zeng2024ssvepgan}. DenoFlow instead learns the path between them: a field network predicts the velocity of the straight path connecting $\tilde{X}'$ to $\tilde{X}$, following the rectified-flow formulation \cite{liu2023rectified,lipman2023flow}. The contaminated input $\tilde{X}'$ serves both as the conditioning input at each point along the path and as the point from which integration starts, so the model never has to generate a trial from noise but only has to move an observation it is given. Denoising is then the forward integration of the predicted field. The network is an encoder-decoder with skip connections, and a category classifier supervises the integrated output so that stimulus-related structure is preserved rather than merely smoothed away. The DenoFlow training optimizes the regression onto the velocity target.

\subsection{Physiological Artifact Injection}\label{sec:artifact}

\begin{figure*}[htpb]
	\subfigure[]{\scalebox{1}[0.9]{\includegraphics[width=0.32\linewidth,clip]{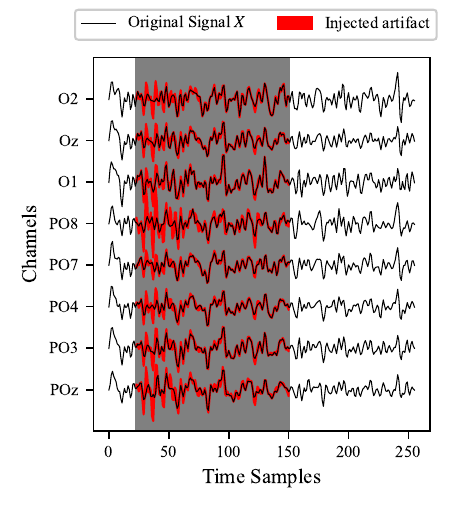}}\label{fig:inj_emg}}
	\subfigure[]{\scalebox{1}[0.9]{\includegraphics[width=0.32\linewidth,clip]{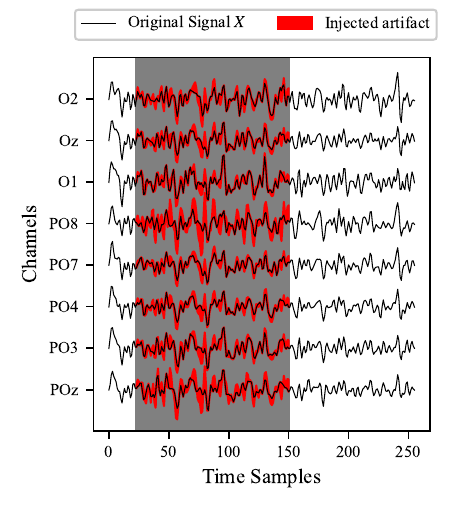}\label{fig:inj_eog}}}
	\subfigure[]{\scalebox{1}[0.9]{\includegraphics[width=0.32\linewidth,clip]{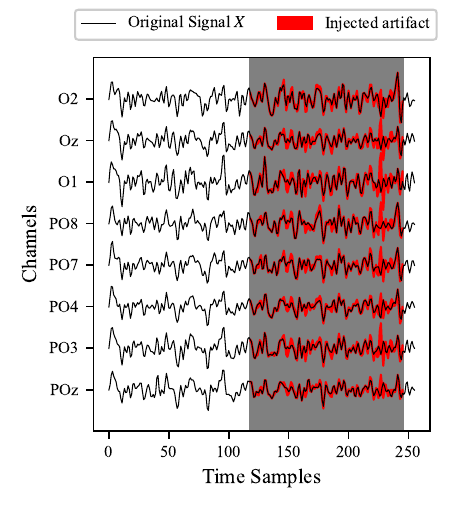}\label{fig:inj_both}}}
	\caption{Physiological artifact injection under the (a) EMG, (b) EOG, and (c) EMG-EOG mixed conditions. Black is the uncontaminated trial $X$, red the injected artifact, and the shaded band is the mask $M$.}
	\label{fig:injection}
\end{figure*}

Training an effective denoiser requires the paired recordings and perturbations. In practice, we manually inject the known perturbations to EEG recordings. Physiological artifacts are non-stationary, spatially structured, and occupy specific frequency bands, so we inject recorded artifacts rather than synthetic perturbations. Segments are drawn from the EMG and EOG libraries of EEGdenoiseNet \cite{zhang2021eegdenoisenet} and resampled to the sampling rate of the target
dataset. Three conditions are considered: EMG only, EOG only, and EMG-EOG mixed. Figure~\ref{fig:injection} shows the three types of physiological artifacts.

Let $X\in\mathbb{R}^{N_e\times T}$ be an uncontaminated trial recorded from $N_e$ electrodes and
$s_k\in\mathbb{R}^{T}$ the $k$-th artifact segment. Each segment is spread over the electrodes by
a log-normal gain drawn afresh for that segment, $g_k^{(c)}=\exp(\varepsilon_{k,c})$ with
$\varepsilon_{k,c}\sim\mathcal{N}(0,\sigma^2)$. Collecting these over the electrodes into
$g_k\in\mathbb{R}^{N_e}$ gives a multi-channel artifact
$A_k=g_k s_k^{\top}$, a coarse stand-in for volume conduction. A binary mask $M$ selects the contaminated region, either a contiguous time
span across all electrodes or a subset of electrodes across the whole trial. Each segment is then
scaled so that within the mask it meets a target signal-to-noise ratio $\gamma$,
\begin{equation}\label{eq:snrscale}
	\alpha_k=\sqrt{\frac{\mathbb{E}\!\left[(X\odot M)^2\right]}
	{\mathbb{E}\!\left[(A_k\odot M)^2\right]\cdot 10^{\gamma/10}}},
\end{equation}
and the contaminated trial is
\begin{equation}\label{eq:inject}
	X'=X+\Big(\textstyle\sum_{k}\alpha_k A_k\Big)\odot M .
\end{equation}
Each segment is scaled on its own, so with $K$ segments the ratio realized over the masked region
is about $\gamma-10\log_{10}K$: the EMG and EOG conditions draw a single segment and sit at
$\gamma$, whereas the mixed condition draws one from each library and is therefore some
$3$~dB harsher. Contamination severity is controlled by the mask fraction and $\gamma$, and the
specific settings used are given in Section~\ref{sect:er}.

\subsection{Network Architecture}

Table~\ref{tab:denonet_network} summarizes the architecture. DenoFlow has two components: a
field network that predicts the velocity of the transport path, and a classifier that supervises
the integrated output.

\begin{table*}[!t]
	\centering
	\caption{DenoFlow architecture. $N_c=N_eN_b$ is the electrode count times the number of filter-bank sub-bands.}
	\centering\setlength{\tabcolsep}{2mm}
	\footnotesize
	\renewcommand{\arraystretch}{0.85}
	\begin{tabular}{c|c|lll}
		\toprule
		Branch & Stage & Layer (Options) & Kernels / Size & Output Shape \\
		\midrule
		\multirow{14}{*}{\shortstack{Field\\Network}}
		& \shortstack{Spatial Conv. 1}
		& Spatial Conv2D ($In{=}(x_t,\tilde{X}')$), BN-PReLU $(\rightarrow x_1)$ & $4N_c$ / $(3,1)$ & $(B, 4N_c, N_c/3, T)$ \\
		\cmidrule{2-5}
		& \shortstack{Temporal Conv. 1}
		& Temporal Conv2D, $stride{=}(1,2)$, BN-PReLU $(\rightarrow x_2)$ & $8N_c$ / $(1,16)$ & $(B, 8N_c, N_c/3, T/2)$ \\
		\cmidrule{2-5}
		& \shortstack{Spatial Conv. 2}
		& Spatial Conv2D, BN-PReLU $(\rightarrow x_3)$ & $8N_c$ / $(N_c/3,1)$ & $(B, 8N_c, 1, T/2)$ \\
		\cmidrule{2-5}
		& \shortstack{Temporal Conv. 2}
		& Temporal Conv2D, $stride{=}(1,2)$, BN-PReLU $(\rightarrow x_4)$ & $16N_c$ / $(1,16)$ & $(B, 16N_c, 1, T/4)$ \\
		\cmidrule{2-5}
		& \multirow{2}{*}{\shortstack{Path\\embedding}}
		& Sinusoidal Encoding ($In{=}t$) & -- & $(B, 16N_c)$ \\
		& & Linear-SiLU-Linear & $16N_c$ / -- & $(B, 16N_c)$ \\
		\cmidrule{2-5}
		& \multirow{3}{*}{\shortstack{Transformer\\Encoder}}
		& Squeeze \& Permute, $dims{=}(0,2,1)$ & -- & $(B, T/4, 16N_c)$ \\
		& & Positional Encoding $+$ path embedding & -- & $(B, T/4, 16N_c)$ \\
		& & Multi-Head Attention, 4 layers, 12 heads & -- & $(B, T/4, 16N_c)$ \\
		\cmidrule{2-5}
		& \shortstack{Temporal Deconv. 2}
		& Skip $(x_4)$; Temporal Deconv2D, $stride{=}(1,2)$ & $8N_c$ / $(1,16)$ & $(B, 8N_c, 1, T/2)$ \\
		\cmidrule{2-5}
		& \shortstack{Spatial Deconv. 2}
		& Skip $(x_3)$; Spatial Deconv2D & $8N_c$ / $(N_c/3,1)$ & $(B, 8N_c, N_c/3, T/2)$ \\
		\cmidrule{2-5}
		& \shortstack{Temporal Deconv. 1}
		& Skip $(x_2)$; Temporal Deconv2D, $stride{=}(1,2)$ & $4N_c$ / $(1,16)$ & $(B, 4N_c, N_c/3, T)$ \\
		\cmidrule{2-5}
		& \shortstack{Spatial Deconv. 1}
		& Skip $(x_1)$; Spatial Deconv2D & $4N_c$ / $(3,1)$ & $(B, 4N_c, N_c, T)$ \\
		\cmidrule{2-5}
		& Output & Conv2D $(\rightarrow f_\theta)$ & $1$ / $(1,1)$ & $(B, 1, N_c, T)$ \\
		\midrule
		\multirow{9}{*}{Classifier}
		& \multirow{2}{*}{\shortstack{Spatial\\Conv.}}
		& Separable Conv2D ($In{=}\hat{X}$ or $\tilde{X}$) & $2N_c$ / $(N_c, 1)$ & $(B, 2N_c, 1, T)$ \\
		& & BN2D-PReLU + Dropout, $p{=}0.5$ & -- & $(B, 2N_c, 1, T)$ \\
		\cmidrule{2-5}
		& \multirow{2}{*}{\shortstack{Temporal\\Conv.}}
		& Temporal Conv2D, $stride{=}(1,2)$ & $4N_c$ / $(1, 10)$ & $(B, 4N_c, 1, T/2)$ \\
		& & BN2D-PReLU + Dropout, $p{=}0.5$ & -- & $(B, 4N_c, 1, T/2)$ \\
		\cmidrule{2-5}
		& \multirow{3}{*}{\shortstack{LSTM\\Block}}
		& Squeeze \& Permute, $dims{=}(0,2,1)$ & -- & $(B, T/2, 4N_c)$ \\
		& & Bi-LSTM, 1 layer & -- & $(B, T/2, 8N_c)$ \\
		& & Flatten, $D_0{=}4N_c{\cdot}T$ & -- & $(B, D_0)$ \\
		\cmidrule{2-5}
		& \multirow{2}{*}{\shortstack{Dense\\Layers}}
		& Linear-PReLU $\times 2$, $D_1{=}D_0{/}{/}10$, $D_2{=}D_1{/}{/}5$ & $D_2$ / -- & $(B, D_2)$ \\
		& & Linear $(\rightarrow x_5)$ $+$ Softmax, $dim{=}1$ & $N_f$ / -- & $(B, N_f)$ \\
		\bottomrule
	\end{tabular}
	\label{tab:denonet_network}
\end{table*}

\subsubsection{Field Network}
The field network $f_\theta$ takes three arguments: the current point $x_t$ on the path, the
contaminated input $\tilde{X}'$ that conditions it, and the path position $t$. The first two are
stacked as two planes of a $(B,2,N_c,T)$ tensor, so conditioning is available at every layer
rather than only at the bottleneck. Here $N_c=N_eN_b$ is the number of rows of that tensor, the
$N_e$ electrodes times the $N_b$ filter-bank sub-bands, which are stacked along the same axis.

The encoder alternates spatial and temporal convolutions. A first spatial convolution reduces the
row axis by the number of sub-bands, a strided temporal convolution then halves the time resolution, the second spatial convolution collapses the remaining rows, and the second strided temporal convolution halves the time axis again. The maps $x_1$ to $x_4$ are kept for skip
connections. A 4-layer, 12-head Transformer encoder with positional encoding then models long-range temporal structure at the bottleneck. The timestep $t$ is mapped by a sinusoidal embedding and a two-layer MLP and added to the Transformer input. Placing it at the bottleneck rather than at the input keeps the conditioning global and describes the whole trial rather than any part of it.

The decoder mirrors the encoder, undoing the four stages in reverse with transposed convolutions and a skip connection at each scale, so that both the electrode structure and the fine temporal detail survive the bottleneck. A final $1\times1$ convolution restores the shape $(B,1,N_c,T)$. The output is a velocity field, not a denoised trial, and the denoised trial is
obtained by integrating it.

\subsubsection{Classifier}
The classifier $C$ shares the model architecture of SSVEPNet \cite{pan2022ssvepnet}. The spatial convolution, temporal convolution, and a Bi-LSTM map a trial to a compact representation, which fully connected layers project to an embedding $x_5$, and a linear head predicts the stimulus class. Spectral normalization is applied to the convolutional and linear layers, and a max-norm constraint is applied to the spatial kernels, bounding the magnitude of the gradient that the classification loss sends back into $f_\theta$.

The classifier is trained on the reference trials and the integrated outputs, and supplies the classification loss. Denoising at test time requires only the field network, and the classifier is used only for training.

\subsection{Network Optimization}\label{sec:objective}

\subsubsection{Velocity Regression}
For a pair $(\tilde{X}',\tilde{X})$, the straight path between them is parameterized by
$t\in[0,1]$,
\begin{equation}\label{eq:path}
	x_t=(1-t)\,\tilde{X}'+t\,\tilde{X},
\end{equation}
and its velocity $v=\tilde{X}-\tilde{X}'$ is constant along the path. The field network regresses that velocity at a position drawn uniformly at random,
\begin{equation}\label{eq:flow}
	\mathcal{L}_{\text{flow}}
	=\mathbb{E}_{t\sim\mathcal{U}(0,1),\,(\tilde{X},\tilde{X}')}
	\frac{1}{N_cT}\left\lVert f_\theta\!\left(x_t,\tilde{X}',t\right)-\left(\tilde{X}-\tilde{X}'\right)\right\rVert_F^2 ,
\end{equation}
where $\lVert\cdot\rVert_F$ is the Frobenius norm. The factor $1/(N_cT)$ turns the squared norm into a mean over the entries of a trial, which keeps the loss terms on a comparable scale. Because $t$ is resampled at each step, the network is supervised at every point of the path rather than only at its start.

\subsubsection{Denoising by Integration}
Denoising integrates the learned field forward from the observation with $S$ Euler steps,
\begin{equation}\label{eq:sample}
	\hat{X}^{(s+1)}=\hat{X}^{(s)}+\tfrac{1}{S}\,
	f_\theta\!\left(\hat{X}^{(s)},\tilde{X}',\tfrac{s}{S}\right),
	\qquad \hat{X}^{(0)}=\tilde{X}',
\end{equation}
giving $\hat{X}=\hat{X}^{(S)}$. At $S=1$, this adds a single predicted residual to the observation, and a larger $S$ re-evaluates the field along the way. Training runs the same recursion with $S_{\text{tr}}$ steps to form the $\hat{X}$ that the reconstruction and classification losses act on. The two counts are independent, so $S$ can be changed after
training without touching the field. The used values are introduced in Section~\ref{sec:params}.

\subsubsection{Signal Reconstruction Regularization}
$\mathcal{L}_{\text{flow}}$ supervises the velocity field at every position along the path,
but places no constraint on the endpoint that the Euler recursion of Eq.~\eqref{eq:sample}
actually produces. That endpoint is the signal a downstream decoder receives, so it is
constrained directly by a reconstruction loss,
\begin{equation}\label{eq:rec}
	\mathcal{L}_{\text{rec}}=\frac{1}{N_cT}\left\lVert\hat{X}-\tilde{X}\right\rVert_F^2 ,
\end{equation}
where $\hat{X}$ is obtained from Eq.~\eqref{eq:sample} with $S_{\text{tr}}$ steps. Since the field is re-evaluated at each step, the gradient propagates through the entire
roll-out, so integration error is attributed to the step at which it arises.

\subsubsection{Overall Objective}
The overall training objective includes the flow matching velocity regression, the reconstruction loss, and the supervised classification loss.

\begin{equation}\label{eq:total}
	\mathcal{L}=\mathcal{L}_{\text{cls}}+\lambda_{\text{flow}}\mathcal{L}_{\text{flow}}
	+\lambda_{\text{rec}}\mathcal{L}_{\text{rec}},
\end{equation}
where $\mathcal{L}_{\text{cls}}=\mathrm{CE}(C(\hat{X}),\,y)$ is the classification loss and both trade-off weights are set to one unless stated otherwise. The classifier is trained alongside the field network on both the reference and the integrated output,
\begin{equation}\label{eq:cls}
	\mathcal{L}_{C}=\mathrm{CE}\!\left(C(\tilde{X}),\,y\right)
	+\mathrm{CE}\!\left(C\!\left(\mathrm{sg}[\hat{X}]\right),\,y\right),
\end{equation}
where $\mathrm{CE}(\cdot)$ is the cross-entropy loss, $y$ the stimulus label, and $\mathrm{sg}[\cdot]$ stops the gradient. The field network $f_\theta$ and classifier $C$ are updated alternately within a mini-batch, $C$ first and then $f_\theta$.

\subsubsection{Training Procedure}
The pseudo-code of DenoFlow is given in Algorithm~\ref{alg:denonet}. Each mini-batch draws a fresh $t$ and a fresh artifact segment, so neither the path position nor the contamination is fixed across epochs.

\begin{algorithm}[htpb]
	\caption{Training and denoising with DenoFlow.}
	\small
	\label{alg:denonet}
	\begin{algorithmic}[1]
		\REQUIRE Training set $\mathcal{T}=\{(X,y)\}$, epochs $E$, training Euler steps $S_{\text{tr}}$
		\ENSURE Trained field network $f_\theta$; denoised trial $\hat{X}$
		\STATE \textit{// Training}
		\FOR{epoch $=1$ to $E$}
		\FOR{each mini-batch $(X,y)$}
		\STATE Inject artifacts: $X' \leftarrow \text{Inject}(X)$ by \cref{eq:inject}
		\STATE Apply the filter bank to obtain $\tilde{X}$ and $\tilde{X}'$
		\STATE Sample $t\sim\mathcal{U}(0,1)$ and form $x_t$ by \cref{eq:path}
		\STATE Compute $\mathcal{L}_{\text{flow}}$ by \cref{eq:flow}
		\STATE Integrate $S_{\text{tr}}$ steps to obtain $\hat{X}$ by \cref{eq:sample}
		\STATE Update $C$ by minimizing $\mathcal{L}_{C}$ in \cref{eq:cls}
		\STATE Update $f_\theta$ by minimizing $\mathcal{L}$ in \cref{eq:total}
		\ENDFOR
		\ENDFOR
		\STATE \textit{// Denoising with the trained $f_\theta$ and $S$ Euler steps}
		\STATE Apply the filter bank to the contaminated trial to obtain $\tilde{X}'$
		\STATE $\hat{X}^{(0)} \leftarrow \tilde{X}'$
		\FOR{$s=0$ to $S-1$}
		\STATE $\hat{X}^{(s+1)} \leftarrow \hat{X}^{(s)} + \tfrac{1}{S}\,f_\theta(\hat{X}^{(s)},\tilde{X}',s/S)$ by \cref{eq:sample}
		\ENDFOR
		\RETURN $f_\theta$, $\hat{X}=\hat{X}^{(S)}$
	\end{algorithmic}
\end{algorithm}

\section{Experiments and Results}\label{sect:er}
Code for DenoFlow, along with all compared baseline models, is available on GitHub\footnote{https://github.com/wzwvv/DenoFlow}.

\subsection{Datasets}
We conducted experiments on two public SSVEP datasets, summarized in Table~\ref{tab:dataset_info}.

\begin{table}[htpb]
	\centering \setlength{\tabcolsep}{0.6mm}
	\renewcommand{\arraystretch}{1}
	\caption{Summary of the two SSVEP datasets.}
	\label{tab:dataset_info}
	\footnotesize
	\begin{tabular}{c|c|c|c|c|c|c}
		\toprule
		\multirow{2}{*}{Dataset} & \multirow{2}{*}{\# Subjects} & \# EEG & Sampling & Trial & \multirow{2}{*}{\# Trials} & \multirow{2}{*}{\# Stimuli} \\
		& & Channels & Rate (Hz) & Length (s) & &\\
		\midrule
		Nakanishi2015 & 10 & 8 & 256 & 1 & 1,800 & 12 \\
		Benchmark & 35 & 9 & 250 & 1 & 8,400 & 40 \\
		\bottomrule
	\end{tabular}
\end{table}

For the Nakanishi2015 SSVEP dataset \cite{nakanishi2015dataset}, 12 target stimuli were displayed on a 27-inch LCD monitor, each flickering at a distinct frequency ranging from 9.25~Hz to 14.75~Hz with a frequency interval of 0.5~Hz. Ten subjects with normal or corrected-to-normal vision participated in the experiment. EEG signals were recorded using a BioSemi ActiveTwo system from 8 electrodes located over the occipital region. The experiment consisted of 15 blocks, each containing 12 trials corresponding to the 12 stimulation frequencies. The EEG data were originally sampled at 2048~Hz and subsequently downsampled to 256~Hz for further analysis.

For the Benchmark SSVEP dataset~\cite{wang2016benchmark}, 64-channel EEG trials were recorded from 35 healthy subjects, including 8 experienced and 27
naive participants. The experiment followed a cue-guided target-selection paradigm using a virtual keyboard composed of 40 visual stimuli. The stimulation frequencies ranged from 8.0~Hz to 15.8~Hz with a step size of 0.2~Hz, and the phase difference between adjacent frequencies was fixed at $0.5\pi$. Six blocks, each containing 40 trials, were collected from each subject using a SynAmps2 system at a sampling rate of 1000~Hz, later downsampled to 250~Hz. EEG signals from nine posterior channels \{Pz, PO3, PO5, PO4, PO6, POz, O1, Oz, and O2\} were selected for analysis, as in \cite{ravi2020CCNN}.

\subsection{Algorithms}
Five decoders were employed for comparison, two traditional algorithms and three deep learning models. Each of them decodes the denoised trials, as well as the uncontaminated reference and the contaminated trial left untouched.
\begin{enumerate}
	\item FBCCA \cite{chen2015fbcca} is a widely used SSVEP decoding algorithm that extends standard CCA~\cite{lin2006cca} by incorporating a filter bank. EEG signals are decomposed into multiple sub-bands, and CCA is performed independently in each sub-band to measure the correlation between EEG responses and predefined sinusoidal reference signals. The final decision is obtained by weighted fusion of correlation coefficients across sub-bands.
	\item eTRCA \cite{nakanishi2017trca} is a supervised spatial filtering approach that maximizes task-related reproducibility across multiple trials. By learning discriminative spatial filters for each stimulus frequency and integrating them into an ensemble framework, eTRCA enhances robustness against noise and inter-trial variability.
	\item EEGNet \cite{lawhern2018eegnet} is a compact CNN specifically designed for EEG decoding. It employs depthwise and separable convolutions to efficiently learn spatial and temporal features with a small number of trainable parameters.
	\item CCNN \cite{ravi2020CCNN} consists of CNN layers designed for spatial and spectral feature extraction, followed by fully connected layers for classification. EEG signals are transformed into the frequency domain by fast Fourier transform, and the real and imaginary components of the complex spectrum are concatenated to form the input feature maps.
	\item SSVEPNet \cite{pan2022ssvepnet} adopts a hybrid architecture that combines one-dimensional CNN layers with an LSTM module. CNN layers extract local temporal features, while the LSTM layer captures long-range temporal dependencies. In addition, spectral normalization is applied to stabilize training, and label smoothing is employed to enhance generalization.
\end{enumerate}

Seven representative denoising models were reproduced and compared with DenoFlow, retrained under the same protocol.
\begin{enumerate}
	\item ResCNN \cite{sun2020novel}: A one-dimensional residual convolutional network operating on the raw waveform, in which parallel residual blocks of different kernel widths are concatenated before projection.
	\item FCNN \cite{zhang2021eegdenoisenet}: A fully connected network applied to the flattened trial, the simplest of the reference architectures benchmarked with EEGdenoiseNet.
	\item SimpleCNN \cite{zhang2021eegdenoisenet}: A shallow stack of one-dimensional convolutions with average pooling, also from that benchmark.
	\item EEGDNet \cite{pu2022eegdnet}: A Transformer-based EEG denoising network that exploits both global and local self-similarity. It uses self-attention blocks to capture long-range dependencies and feed-forward blocks to model local structure, enabling end-to-end artifact suppression with high efficiency.
	\item GCTNet \cite{yin2023gan}: A GAN-guided parallel CNN-Transformer network for EEG artifact removal. The generator combines CNN and Transformer blocks to learn local and global temporal features, while the discriminator enforces distribution alignment between denoised and clean EEG signals.
	\item TOED-GAN \cite{zeng2024ssvepgan}: A task-oriented GAN-based denoising network for SSVEP, and the only baseline designed for this paradigm. Its generator is a U-Net with a bidirectional LSTM bottleneck, and its discriminator is an SSVEPNet backbone with parallel authenticity and classification heads. It is trained with a class-aware loss and an adversarial loss, and carries no reconstruction term. We reproduced it with the published architecture and hyperparameters.
	\item CT-DCENet \cite{tang2025ct}: A dual-stage CNN-Transformer ensemble model for EEG denoising. It first ensembles multiple learners for coarse reconstruction of low-frequency morphology and then refines high-frequency details from residual information.
\end{enumerate}

TOED-GAN is designed for multi-channel SSVEP trials, while the remaining models were developed for single-channel artifact removal and are reproduced here to filter-bank multi-channel input.

\subsection{Implementation}
\subsubsection{Evaluation Tasks}
All experiments are conducted under within-subject settings following the chronological setting in \cite{wang2025dbconformer}, where the training and test data are drawn chronologically from the same subject. The contaminated trial $\tilde{X}'$ is passed through the denoiser under evaluation and the output $\hat{X}$ is decoded: signal-level fidelity is measured against the reference $\tilde{X}$, and decoding accuracy is measured on $\hat{X}$.

\subsubsection{Evaluation Metrics}
The denoising performance is assessed at the signal and task levels. At the signal level, we report the correlation coefficient (CC) between the filtered $\hat{X}$ and the denoised $\tilde{X}$, the relative root-mean-square error (RRMSE), which normalizes the error by the power of the reference, and the signal-to-noise ratio improvement (SNRI), the ratio measured on the denoised trial minus the one measured on the contaminated trial, so that a negative SNRI means the denoiser has left the trial noisier than it found it. At the task level we report classification accuracy, the fraction of correctly identified trials. We reported the mean and standard deviation over subjects. The best average result of each artifact type is marked in bold and the second best by an underlined.

\subsubsection{Parameter Settings}\label{sec:params}
Contamination is injected in the temporal mode under three artifact conditions, EMG only, EOG only and EMG-EOG mixed, as described in Section~\ref{sec:artifact}. Two factors set its severity: the mask fraction and the target ratio $\gamma$. Unless stated otherwise, results are reported at the reference condition, a mask fraction of $0.5$ and $\gamma=0$~dB. Section~\ref{sec:noise} varies both factors. One segment is drawn per artifact library, so the EMG and EOG conditions inject a single segment and the mixed condition two, and the channel-gain spread of Section~\ref{sec:artifact} is $\sigma=0.3$. Note that $\gamma$ is set before the filter bank, so the ratio realized at the network input is higher than the nominal one.

The field network and the classifier are optimized with Adam at a learning rate of $1\times10^{-3}$, weight decay $1\times10^{-4}$, batch size $36$, and $300$ epochs. The objective of Eq.~\eqref{eq:total} integrates $S_{\text{tr}}=4$ steps and denoising with $S=4$. The trade-off weights $\lambda_{\text{flow}}$ and $\lambda_{\text{rec}}$ are set to $1$.

Five-fold cross-validation is adopted for Nakanishi2015, and six-fold cross-validation for Benchmark, since the number of trials per class per subject is 6 in Benchmark. For both datasets, a filter bank of 3 sub-bands is used, each a Chebyshev type-I band-pass filter with $0.5$~dB passband ripple and $40$~dB stopband attenuation, applied with zero-phase forward-backward filtering. Each filtered sub-band passes from its lower edge to $80$~Hz. The lower edges are $9$, $18$ and $27$~Hz on Nakanishi2015, and $8$, $16$ and $24$~Hz on Benchmark, each set just below the lowest stimulus frequency of its dataset. Stacking the sub-bands along the channel dimension yields $N_c$ rows for the network input.

\subsection{Denoising Results}
\subsubsection{Signal Fidelity}
Tables~\ref{tab:fid_ds1} and~\ref{tab:fid_ds2} report how closely the denoised trial follows the reference on Nakanishi2015 and Benchmark. Three points stand out:
\begin{enumerate}
	\item DenoFlow ranks the first on all metrics under all artifact types on both datasets.
	EEGDNet obtained the second-best.
	\item The CC of DenoFlow stayed above $0.86$ in each panel, and this lead widened on the harder Benchmark dataset.
	\item Under the EOG artifact, the baselines lowered the SNR, yielding $-5.88$~dB on Nakanishi2015, whereas DenoFlow was near zero there and is the only method to raise the ratio on Benchmark. Ocular artifacts overlap the stimulus band, so a denoiser with no model of the response removes signal along with the artifact.
\end{enumerate}

To evaluate the robustness of DenoFlow, we conducted two-sided Wilcoxon signed-rank
tests \cite{wilcoxon1945} against each of the seven baselines. Samples were paired over subjects,
giving 10 pairs on Nakanishi2015 and 35 on Benchmark, and the $p$-values were adjusted by the
Holm-Bonferroni procedure \cite{holm1979} within each panel of seven baselines. The two datasets
reach different significance levels because of sample size rather than effect size. With 10
subjects, an exact signed-rank test cannot return a two-sided $p$ below $1.95\times10^{-3}$, or
$1.37\times10^{-2}$ after correction, so a comparison won on all 10 subjects still carries a
single asterisk.

\begin{table}[!t]
	\centering
	\caption{Signal-level fidelity on Nakanishi2015. Higher is better for CC and SNRI, lower for
	RRMSE. Statistical significance is denoted by * ($p<0.05$), ** ($p<0.005$), and *** ($p<0.001$).}
	\label{tab:fid_ds1}
	\footnotesize
	\setlength{\tabcolsep}{3pt}
	\begin{tabular}{@{}l|l|ccc@{}}
		\toprule
		Artifact & Method & CC & RRMSE & SNRI (dB) \\
		\midrule
		\multirow{8}{*}{EMG} & ResCNN & 0.7980$\pm$0.0384$^{*}$ & 0.5933$\pm$0.0459$^{*}$ & 0.61$\pm$0.89$^{*}$ \\
		 & FCNN & 0.4901$\pm$0.0720$^{*}$ & 0.8598$\pm$0.0440$^{*}$ & $-$2.83$\pm$1.16$^{*}$ \\
		 & SimpleCNN & 0.8084$\pm$0.0281$^{*}$ & 0.5906$\pm$0.0396$^{*}$ & 0.69$\pm$0.78$^{*}$ \\
		 & EEGDNet & \underline{0.8396$\pm$0.0201}$^{*}$ & \underline{0.5432$\pm$0.0277}$^{*}$ & \underline{1.38$\pm$0.71}$^{*}$ \\
		 & GCTNet & 0.7986$\pm$0.0197$^{*}$ & 0.5997$\pm$0.0239$^{*}$ & 0.40$\pm$0.81$^{*}$ \\
		 & TOED-GAN & 0.7133$\pm$0.0287$^{*}$ & 0.7323$\pm$0.0452$^{*}$ & $-$1.39$\pm$0.62$^{*}$ \\
		 & CT-DCENet & 0.5229$\pm$0.1794$^{*}$ & 0.9209$\pm$0.0490$^{*}$ & $-$3.47$\pm$0.99$^{*}$ \\
		 & \textbf{DenoFlow} & \textbf{0.9099$\pm$0.0100} & \textbf{0.3989$\pm$0.0219} & \textbf{4.22$\pm$1.11} \\
		\midrule
		\multirow{8}{*}{EOG} & ResCNN & 0.9390$\pm$0.0295$^{*}$ & 0.3411$\pm$0.0765$^{*}$ & $-$8.38$\pm$2.13$^{*}$ \\
		 & FCNN & 0.5491$\pm$0.0677$^{*}$ & 0.8285$\pm$0.0468$^{*}$ & $-$16.47$\pm$2.04$^{*}$ \\
		 & SimpleCNN & 0.9526$\pm$0.0197$^{*}$ & 0.2964$\pm$0.0590$^{*}$ & $-$7.10$\pm$1.99$^{*}$ \\
		 & EEGDNet & \underline{0.9668$\pm$0.0109}$^{*}$ & \underline{0.2586$\pm$0.0343}$^{*}$ & \underline{$-$5.88$\pm$1.23}$^{*}$ \\
		 & GCTNet & 0.9236$\pm$0.0129$^{*}$ & 0.4109$\pm$0.0283$^{*}$ & $-$10.30$\pm$1.65$^{*}$ \\
		 & TOED-GAN & 0.8366$\pm$0.0393$^{*}$ & 0.6258$\pm$0.0758$^{*}$ & $-$13.90$\pm$1.49$^{*}$ \\
		 & CT-DCENet & 0.6877$\pm$0.1788$^{*}$ & 0.8623$\pm$0.0590$^{*}$ & $-$16.86$\pm$1.93$^{*}$ \\
		 & \textbf{DenoFlow} & \textbf{0.9815$\pm$0.0081} & \textbf{0.1597$\pm$0.0333} & \textbf{$-$0.16$\pm$0.37} \\
		\midrule
		\multirow{8}{*}{Mixed} & ResCNN & 0.7812$\pm$0.0413$^{*}$ & 0.6191$\pm$0.0460$^{*}$ & 0.79$\pm$0.91$^{*}$ \\
		 & FCNN & 0.4803$\pm$0.0719$^{*}$ & 0.8658$\pm$0.0431$^{*}$ & $-$2.30$\pm$1.19$^{*}$ \\
		 & SimpleCNN & 0.7891$\pm$0.0320$^{*}$ & 0.6225$\pm$0.0416$^{*}$ & 0.78$\pm$0.82$^{*}$ \\
		 & EEGDNet & \underline{0.8245$\pm$0.0215}$^{*}$ & \underline{0.5699$\pm$0.0291}$^{*}$ & \underline{1.53$\pm$0.67}$^{*}$ \\
		 & GCTNet & 0.7844$\pm$0.0193$^{*}$ & 0.6195$\pm$0.0221$^{*}$ & 0.68$\pm$0.83$^{*}$ \\
		 & TOED-GAN & 0.7010$\pm$0.0282$^{*}$ & 0.7433$\pm$0.0404$^{*}$ & $-$0.94$\pm$0.69$^{*}$ \\
		 & CT-DCENet & 0.5936$\pm$0.1848$^{*}$ & 0.9036$\pm$0.0467$^{*}$ & $-$2.72$\pm$1.05$^{*}$ \\
		 & \textbf{DenoFlow} & \textbf{0.8947$\pm$0.0087} & \textbf{0.4339$\pm$0.0161} & \textbf{3.99$\pm$0.92} \\
		\bottomrule
	\end{tabular}
\end{table}

\begin{table}[!t]
	\centering
	\caption{Signal-level fidelity on Benchmark. Higher is better for CC and SNRI, lower for
	RRMSE. Statistical significance is denoted by * ($p<0.05$), ** ($p<0.005$), and *** ($p<0.001$).}
	\label{tab:fid_ds2}
	\footnotesize
	\setlength{\tabcolsep}{3pt}
	\begin{tabular}{@{}l|l|ccc@{}}
		\toprule
		Artifact & Method & CC & RRMSE & SNRI (dB) \\
		\midrule
		\multirow{8}{*}{EMG} & ResCNN & 0.6239$\pm$0.0717$^{***}$ & 0.8746$\pm$0.1180$^{***}$ & 3.78$\pm$2.52$^{***}$ \\
		 & FCNN & 0.3707$\pm$0.0811$^{***}$ & 0.9811$\pm$0.1652$^{***}$ & 2.50$\pm$2.44$^{***}$ \\
		 & SimpleCNN & 0.6257$\pm$0.0722$^{***}$ & 0.9208$\pm$0.1519$^{***}$ & 3.48$\pm$2.38$^{***}$ \\
		 & EEGDNet & \underline{0.7017$\pm$0.0721}$^{***}$ & \underline{0.7175$\pm$0.0909}$^{***}$ & \underline{5.29$\pm$2.28}$^{***}$ \\
		 & GCTNet & 0.6687$\pm$0.0596$^{***}$ & 0.7544$\pm$0.0749$^{***}$ & 4.76$\pm$2.56$^{***}$ \\
		 & TOED-GAN & 0.5544$\pm$0.1439$^{***}$ & 0.8938$\pm$0.1383$^{***}$ & 3.42$\pm$2.41$^{***}$ \\
		 & CT-DCENet & 0.3846$\pm$0.2257$^{***}$ & 0.9500$\pm$0.0422$^{***}$ & 2.55$\pm$3.10$^{***}$ \\
		 & \textbf{DenoFlow} & \textbf{0.8933$\pm$0.0628} & \textbf{0.4248$\pm$0.1638} & \textbf{10.75$\pm$1.54} \\
		\midrule
		\multirow{8}{*}{EOG} & ResCNN & 0.8768$\pm$0.0467$^{***}$ & 0.4853$\pm$0.0994$^{***}$ & $-$4.20$\pm$2.44$^{***}$ \\
		 & FCNN & 0.5137$\pm$0.0628$^{***}$ & 0.8684$\pm$0.0632$^{***}$ & $-$9.90$\pm$2.88$^{***}$ \\
		 & SimpleCNN & 0.8802$\pm$0.0521$^{***}$ & 0.4807$\pm$0.1198$^{***}$ & $-$3.92$\pm$1.96$^{***}$ \\
		 & EEGDNet & \underline{0.9068$\pm$0.0476}$^{***}$ & \underline{0.4018$\pm$0.0971}$^{***}$ & \underline{$-$2.25$\pm$1.55}$^{***}$ \\
		 & GCTNet & 0.8639$\pm$0.0458$^{***}$ & 0.5237$\pm$0.0855$^{***}$ & $-$5.20$\pm$2.41$^{***}$ \\
		 & TOED-GAN & 0.6886$\pm$0.1407$^{***}$ & 0.7489$\pm$0.0899$^{***}$ & $-$8.36$\pm$2.82$^{***}$ \\
		 & CT-DCENet & 0.5839$\pm$0.2719$^{***}$ & 0.8729$\pm$0.0787$^{***}$ & $-$10.01$\pm$3.13$^{***}$ \\
		 & \textbf{DenoFlow} & \textbf{0.9606$\pm$0.0234} & \textbf{0.2336$\pm$0.0708} & \textbf{3.51$\pm$1.34} \\
		\midrule
		\multirow{8}{*}{Mixed} & ResCNN & 0.5904$\pm$0.0689$^{***}$ & 0.9145$\pm$0.1095$^{***}$ & 4.37$\pm$2.63$^{***}$ \\
		 & FCNN & 0.3468$\pm$0.0811$^{***}$ & 1.0070$\pm$0.1968$^{***}$ & 3.34$\pm$2.31$^{***}$ \\
		 & SimpleCNN & 0.5914$\pm$0.0693$^{***}$ & 0.9659$\pm$0.1455$^{***}$ & 4.03$\pm$2.49$^{***}$ \\
		 & EEGDNet & \underline{0.6646$\pm$0.0694}$^{***}$ & \underline{0.7620$\pm$0.0844}$^{***}$ & \underline{5.74$\pm$2.40}$^{***}$ \\
		 & GCTNet & 0.6372$\pm$0.0581$^{***}$ & 0.7857$\pm$0.0744$^{***}$ & 5.41$\pm$2.61$^{***}$ \\
		 & TOED-GAN & 0.5391$\pm$0.1347$^{***}$ & 0.8949$\pm$0.1291$^{***}$ & 4.42$\pm$2.56$^{***}$ \\
		 & CT-DCENet & 0.3689$\pm$0.1975$^{***}$ & 0.9567$\pm$0.0365$^{***}$ & 3.52$\pm$3.13$^{***}$ \\
		 & \textbf{DenoFlow} & \textbf{0.8636$\pm$0.0778} & \textbf{0.4971$\pm$0.2207} & \textbf{10.41$\pm$1.36} \\
		\bottomrule
	\end{tabular}
\end{table}

\subsubsection{Decoding Accuracy}
Tables~\ref{tab:acc_ds1} and~\ref{tab:acc_ds2} report the accuracies of five decoders on the denoised trials, together with those obtained using the uncontaminated reference trials (Ref.) and the contaminated trials without denoising (Art.). Observe that:
\begin{enumerate}
    \item DenoFlow achieved the highest average accuracy across all artifact types on both datasets. On Benchmark, all comparisons with the competing denoising methods remained statistically significant after Holm correction. Its largest margin over the strongest baseline was observed under mixed EMG-EOG contamination.
    \item DenoFlow was also the only approach that consistently outperformed Art. under each artifact type on both datasets. The improvements were more pronounced on Benchmark, where contamination caused greater accuracy degradation.
    \item The benefit of denoising generally increased with the severity of artifact-induced degradation. For example, under EMG contamination on Benchmark, the accuracy loss relative to Ref. was 22.50\% for EEGNet, compared with 15.71\% for eTRCA, and DenoFlow recovered more of the larger loss. This pattern is consistent with the greater robustness of decoders that aggregate information across trials or rely on band-limited templates. Since they were less affected by contamination, their gains from denoising were correspondingly smaller.
\end{enumerate}

\begin{table*}[!t]
	\centering
	\caption{Average decoding accuracies (\%) on Nakanishi2015. Ref.\ and Art.\ decode the
	uncontaminated reference and the untouched contaminated trial, and do not compete for the
	ranking.}
	\label{tab:acc_ds1}
	\footnotesize
	\setlength{\tabcolsep}{3pt}
	\resizebox{\linewidth}{!}{%
	\begin{tabular}{@{}l|l|cc|cccccccc@{}}
		\toprule
		Artifact & Decoder & Ref. & Art. & ResCNN & FCNN & SimpleCNN & EEGDNet & GCTNet & TOED-GAN & CT-DCENet & \textbf{DenoFlow} \\
		\midrule
		\multirow{6}{*}{EMG} & FBCCA & 83.61$\pm$15.18 & 79.44$\pm$18.34 & 77.78$\pm$19.42 & 63.89$\pm$24.91 & 77.22$\pm$22.10 & 75.28$\pm$21.05 & \textbf{79.72$\pm$17.91} & \underline{78.89$\pm$19.30} & 73.89$\pm$18.67 & \textbf{79.72$\pm$18.00} \\
		 & eTRCA & 98.33$\pm$2.99 & 96.11$\pm$6.31 & 96.67$\pm$5.37 & 90.28$\pm$17.09 & \underline{96.94$\pm$4.62} & 96.67$\pm$5.37 & \underline{96.94$\pm$4.80} & 82.78$\pm$20.86 & 94.44$\pm$6.68 & \textbf{97.78$\pm$3.15} \\
		 & EEGNet & 95.28$\pm$8.59 & 94.72$\pm$8.92 & 92.78$\pm$10.33 & 87.22$\pm$22.35 & \underline{94.17$\pm$9.02} & 93.89$\pm$9.51 & 92.78$\pm$9.99 & 88.06$\pm$14.29 & 90.56$\pm$11.87 & \textbf{95.00$\pm$7.72} \\
		 & CCNN & 94.17$\pm$6.06 & 91.94$\pm$8.92 & 92.22$\pm$6.39 & 85.28$\pm$20.08 & \textbf{93.06$\pm$5.89} & \underline{92.50$\pm$7.64} & \underline{92.50$\pm$6.42} & 88.61$\pm$12.66 & 88.06$\pm$13.48 & \underline{92.50$\pm$7.41} \\
		 & SSVEPNet & 99.72$\pm$0.88 & 99.17$\pm$1.87 & 98.89$\pm$1.94 & 92.22$\pm$17.46 & \underline{99.17$\pm$1.87} & \underline{99.17$\pm$1.34} & 98.33$\pm$2.34 & 95.00$\pm$5.97 & 95.00$\pm$7.38 & \textbf{99.72$\pm$0.88} \\
		\cmidrule(l){2-12}
		 & \textit{Avg.} & 94.22$\pm$5.60 & 92.28$\pm$7.86 & 91.67$\pm$7.86 & 83.78$\pm$19.39 & \underline{92.11$\pm$7.75} & 91.50$\pm$7.32 & 92.06$\pm$7.27 & 86.67$\pm$13.68 & 88.39$\pm$11.04 & \textbf{92.94$\pm$6.64} \\
		\midrule
		\multirow{6}{*}{EOG} & FBCCA & 83.61$\pm$15.18 & 82.22$\pm$16.15 & 81.94$\pm$16.63 & 66.11$\pm$23.45 & \underline{82.22$\pm$16.93} & 80.28$\pm$16.11 & \textbf{82.78$\pm$15.09} & 80.28$\pm$16.64 & 77.78$\pm$22.64 & \underline{82.22$\pm$16.26} \\
		 & eTRCA & 98.33$\pm$2.99 & 98.06$\pm$2.94 & 97.50$\pm$3.81 & 92.78$\pm$15.56 & 97.22$\pm$3.93 & \underline{97.78$\pm$3.15} & 97.22$\pm$4.14 & 86.39$\pm$14.61 & 95.83$\pm$6.45 & \textbf{98.33$\pm$2.99} \\
		 & EEGNet & 95.28$\pm$8.59 & 95.56$\pm$9.64 & \textbf{95.56$\pm$7.08} & 90.56$\pm$18.85 & 94.44$\pm$8.69 & \underline{94.72$\pm$7.80} & \textbf{95.56$\pm$7.08} & 88.06$\pm$10.88 & 92.22$\pm$12.27 & \textbf{95.56$\pm$7.43} \\
		 & CCNN & 94.17$\pm$6.06 & 93.33$\pm$6.70 & \underline{93.61$\pm$5.25} & 88.89$\pm$16.62 & \underline{93.61$\pm$6.29} & \textbf{94.17$\pm$6.47} & 92.78$\pm$5.89 & 91.67$\pm$7.52 & 88.33$\pm$15.54 & \textbf{94.17$\pm$6.06} \\
		 & SSVEPNet & 99.72$\pm$0.88 & 99.72$\pm$0.88 & \textbf{99.72$\pm$0.88} & 93.89$\pm$14.69 & \underline{99.44$\pm$1.17} & \underline{99.44$\pm$1.17} & \underline{99.44$\pm$1.17} & 95.00$\pm$3.88 & 95.00$\pm$7.50 & \textbf{99.72$\pm$0.88} \\
		\cmidrule(l){2-12}
		 & \textit{Avg.} & 94.22$\pm$5.60 & 93.78$\pm$6.14 & \underline{93.67$\pm$5.89} & 86.44$\pm$16.94 & 93.39$\pm$6.42 & 93.28$\pm$5.73 & 93.56$\pm$5.92 & 88.28$\pm$10.18 & 89.83$\pm$11.76 & \textbf{94.00$\pm$5.77} \\
		\midrule
		\multirow{6}{*}{Mixed} & FBCCA & 83.61$\pm$15.18 & 80.28$\pm$17.74 & 77.78$\pm$19.07 & 61.39$\pm$26.53 & 77.22$\pm$18.32 & 76.39$\pm$19.43 & \underline{79.72$\pm$18.10} & 75.28$\pm$19.75 & 71.67$\pm$20.45 & \textbf{82.50$\pm$14.58} \\
		 & eTRCA & 98.33$\pm$2.99 & 98.33$\pm$2.99 & 97.50$\pm$3.33 & 92.78$\pm$12.51 & \underline{98.33$\pm$2.34} & 96.94$\pm$5.15 & 97.50$\pm$3.81 & 85.28$\pm$14.11 & 94.44$\pm$8.07 & \textbf{98.61$\pm$2.70} \\
		 & EEGNet & 95.28$\pm$8.59 & 93.33$\pm$8.61 & \underline{93.89$\pm$9.51} & 86.67$\pm$20.24 & 93.06$\pm$10.98 & 93.33$\pm$9.37 & 92.50$\pm$9.45 & 83.89$\pm$17.21 & 89.17$\pm$14.31 & \textbf{94.44$\pm$9.71} \\
		 & CCNN & 94.17$\pm$6.06 & 92.78$\pm$6.95 & 92.50$\pm$6.01 & 85.00$\pm$19.07 & \underline{92.78$\pm$6.95} & \underline{92.78$\pm$7.08} & \underline{92.78$\pm$6.57} & 88.89$\pm$11.42 & 88.06$\pm$14.70 & \textbf{93.33$\pm$5.27} \\
		 & SSVEPNet & 99.72$\pm$0.88 & 99.17$\pm$1.87 & \textbf{99.72$\pm$0.88} & 93.61$\pm$13.03 & \underline{99.44$\pm$1.76} & \underline{99.44$\pm$1.17} & 98.89$\pm$1.94 & 93.61$\pm$6.42 & 95.28$\pm$6.68 & \textbf{99.72$\pm$0.88} \\
		\cmidrule(l){2-12}
		 & \textit{Avg.} & 94.22$\pm$5.60 & 92.78$\pm$6.48 & \underline{92.28$\pm$6.76} & 83.89$\pm$17.31 & 92.17$\pm$7.01 & 91.78$\pm$7.27 & \underline{92.28$\pm$6.97} & 85.39$\pm$12.89 & 87.72$\pm$12.24 & \textbf{93.72$\pm$5.77} \\
		\bottomrule
	\end{tabular}}
\end{table*}

\begin{table*}[!t]
	\centering
	\caption{Average decoding accuracies (\%) on Benchmark. Ref.\ and Art.\ decode the
	uncontaminated reference and the untouched contaminated trial, and do not compete for the
	ranking.}
	\label{tab:acc_ds2}
	\footnotesize
	\setlength{\tabcolsep}{3pt}
	\resizebox{\linewidth}{!}{%
	\begin{tabular}{@{}l|l|cc|cccccccc@{}}
		\toprule
		Artifact & Decoder & Ref. & Art. & ResCNN & FCNN & SimpleCNN & EEGDNet & GCTNet & TOED-GAN & CT-DCENet & \textbf{DenoFlow} \\
		\midrule
		\multirow{6}{*}{EMG} & FBCCA & 70.86$\pm$25.58 & 64.57$\pm$25.86 & 53.50$\pm$24.89 & 25.57$\pm$21.09 & 53.64$\pm$25.13 & 47.50$\pm$26.44 & \underline{58.50$\pm$25.98} & 51.50$\pm$26.02 & 35.71$\pm$26.67 & \textbf{64.43$\pm$26.60} \\
		 & eTRCA & 93.50$\pm$12.78 & 77.79$\pm$26.88 & 85.00$\pm$21.83 & 60.14$\pm$29.26 & 84.86$\pm$21.45 & 84.00$\pm$21.10 & \underline{88.57$\pm$17.59} & 65.64$\pm$29.89 & 64.86$\pm$33.81 & \textbf{89.21$\pm$19.86} \\
		 & EEGNet & 70.36$\pm$21.70 & 47.86$\pm$22.80 & \underline{57.43$\pm$23.43} & 39.36$\pm$22.81 & \underline{57.43$\pm$23.89} & 54.21$\pm$23.75 & 54.71$\pm$21.95 & 54.14$\pm$26.51 & 47.64$\pm$25.35 & \textbf{64.36$\pm$24.10} \\
		 & CCNN & 70.86$\pm$23.24 & 53.36$\pm$25.84 & 55.21$\pm$24.07 & 36.21$\pm$21.65 & 54.71$\pm$24.34 & 54.79$\pm$25.80 & 58.64$\pm$23.20 & \underline{65.29$\pm$28.66} & 48.43$\pm$27.61 & \textbf{67.43$\pm$24.89} \\
		 & SSVEPNet & 93.00$\pm$16.68 & 79.93$\pm$29.13 & 85.21$\pm$24.01 & 63.07$\pm$27.65 & 85.29$\pm$23.43 & 83.57$\pm$25.05 & \textbf{89.07$\pm$19.52} & 76.43$\pm$29.50 & 67.36$\pm$33.94 & \underline{88.57$\pm$23.44} \\
		\cmidrule(l){2-12}
		 & \textit{Avg.} & 79.71$\pm$18.09 & 64.70$\pm$23.42 & 67.27$\pm$21.75 & 44.87$\pm$23.37 & 67.19$\pm$21.67 & 64.81$\pm$22.30 & \underline{69.90$\pm$19.85} & 62.60$\pm$27.05 & 52.80$\pm$28.20 & \textbf{74.80$\pm$21.62} \\
		\midrule
		\multirow{6}{*}{EOG} & FBCCA & 70.86$\pm$25.58 & 64.29$\pm$25.98 & 60.29$\pm$27.03 & 31.93$\pm$24.06 & 61.07$\pm$27.09 & 60.86$\pm$25.79 & \underline{65.21$\pm$25.35} & 50.50$\pm$27.02 & 48.36$\pm$30.57 & \textbf{67.50$\pm$24.83} \\
		 & eTRCA & 93.50$\pm$12.78 & 88.36$\pm$19.84 & 88.93$\pm$19.11 & 71.43$\pm$27.49 & 89.36$\pm$17.59 & 90.21$\pm$17.40 & \underline{90.36$\pm$16.99} & 62.93$\pm$30.20 & 74.07$\pm$35.41 & \textbf{92.00$\pm$14.46} \\
		 & EEGNet & 70.36$\pm$21.70 & 61.93$\pm$24.40 & \underline{63.50$\pm$23.63} & 49.00$\pm$26.50 & 62.50$\pm$23.13 & 63.14$\pm$23.91 & 61.14$\pm$24.17 & 55.21$\pm$25.34 & 56.86$\pm$28.64 & \textbf{67.93$\pm$22.30} \\
		 & CCNN & 70.86$\pm$23.24 & 63.43$\pm$26.49 & 63.50$\pm$23.26 & 48.43$\pm$25.57 & 63.14$\pm$23.99 & 63.14$\pm$24.07 & 65.57$\pm$22.83 & \underline{66.21$\pm$25.66} & 55.43$\pm$29.09 & \textbf{68.71$\pm$23.00} \\
		 & SSVEPNet & 93.00$\pm$16.68 & 86.71$\pm$26.41 & 88.79$\pm$21.53 & 73.07$\pm$27.83 & 89.36$\pm$21.47 & 89.00$\pm$21.53 & \underline{90.50$\pm$20.32} & 79.00$\pm$28.16 & 76.07$\pm$34.04 & \textbf{90.57$\pm$20.58} \\
		\cmidrule(l){2-12}
		 & \textit{Avg.} & 79.71$\pm$18.09 & 72.94$\pm$22.17 & 73.00$\pm$21.23 & 54.77$\pm$24.74 & 73.09$\pm$20.79 & 73.27$\pm$20.63 & \underline{74.56$\pm$20.19} & 62.77$\pm$25.87 & 62.16$\pm$30.16 & \textbf{77.34$\pm$19.32} \\
		\midrule
		\multirow{6}{*}{Mixed} & FBCCA & 70.86$\pm$25.58 & 59.50$\pm$25.77 & 47.57$\pm$26.74 & 24.21$\pm$18.88 & 48.43$\pm$24.97 & 42.57$\pm$24.78 & \underline{53.00$\pm$24.52} & 49.57$\pm$26.96 & 35.79$\pm$22.84 & \textbf{60.00$\pm$27.31} \\
		 & eTRCA & 93.50$\pm$12.78 & 76.07$\pm$27.56 & 81.93$\pm$23.41 & 57.36$\pm$29.25 & 81.57$\pm$23.31 & 79.93$\pm$22.91 & \underline{85.50$\pm$20.33} & 63.43$\pm$30.09 & 62.57$\pm$33.96 & \textbf{86.57$\pm$21.01} \\
		 & EEGNet & 70.36$\pm$21.70 & 47.93$\pm$24.80 & 52.21$\pm$23.07 & 35.50$\pm$22.58 & 52.29$\pm$24.01 & 50.00$\pm$23.45 & 50.64$\pm$21.39 & \underline{53.07$\pm$25.24} & 47.64$\pm$26.82 & \textbf{62.36$\pm$23.79} \\
		 & CCNN & 70.86$\pm$23.24 & 50.14$\pm$26.41 & 50.50$\pm$23.85 & 31.71$\pm$19.70 & 49.07$\pm$24.85 & 51.64$\pm$24.28 & 55.86$\pm$24.73 & \textbf{65.07$\pm$28.76} & 47.21$\pm$26.48 & \underline{64.36$\pm$26.65} \\
		 & SSVEPNet & 93.00$\pm$16.68 & 77.79$\pm$29.60 & 81.86$\pm$26.28 & 58.50$\pm$29.99 & 81.07$\pm$26.38 & 79.93$\pm$25.46 & \textbf{87.00$\pm$21.38} & 74.14$\pm$29.61 & 66.64$\pm$32.52 & \underline{86.29$\pm$24.42} \\
		\cmidrule(l){2-12}
		 & \textit{Avg.} & 79.71$\pm$18.09 & 62.29$\pm$24.41 & 62.81$\pm$22.62 & 41.46$\pm$22.85 & 62.49$\pm$22.81 & 60.81$\pm$22.28 & \underline{66.40$\pm$20.54} & 61.06$\pm$27.09 & 51.97$\pm$27.54 & \textbf{71.91$\pm$22.63} \\
		\bottomrule
	\end{tabular}}
\end{table*}

\subsection{Effect of Noise Intensity}\label{sec:noise}
\begin{figure*}[htpb]\centering
	\includegraphics[width=0.75\textwidth]{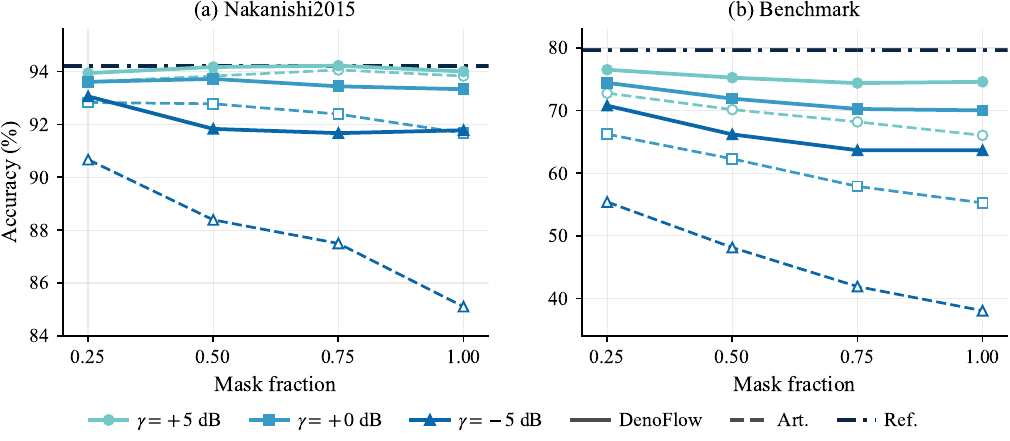}
	\caption{Effect of noise intensity. Denoising flattens the loss: the DenoFlow output degrades far less than the contaminated trial as either factor grows.}
	\label{fig:noise_intensity}
\end{figure*}

The mask fraction and target ratio $\gamma$, which jointly determine contamination severity, were varied in a full factorial design over $\{0.25,0.5,0.75,1.0\}\times\{+5,0,-5\}$~dB. The complete grid was evaluated on both datasets, yielding 12 conditions per dataset. Within each condition, DenoFlow and Art. were evaluated using the same contaminated trials, enabling a direct comparison. Results are illustrated in Figure~\ref{fig:noise_intensity}:
\begin{enumerate}
    \item Contamination became more detrimental as the mask fraction increased or $\gamma$ decreased, with the two factors jointly amplifying the degradation. On Benchmark, the accuracy of Art. decreased from 72.79\% under the mildest condition to 38.04\% under the most severe condition. This decrease was approximately four times that observed on Nakanishi2015. Across the evaluated grid, changes in $\gamma$ had a larger effect than those in the mask fraction, while the latter determined the temporal extent of contamination.
   \item DenoFlow yielded larger absolute improvements as contamination became more severe. On Benchmark, it improved accuracy over Art. by less than 4\% under the mildest condition and by more than 25\% under the most severe condition. Nakanishi2015 exhibited the same trend, although with smaller absolute changes. Thus, the benefit of denoising generally increased with the artifact-induced accuracy loss.
   \item Accuracy after denoising varied substantially less across contamination conditions than that obtained with Art. On Benchmark, the range across the 12 conditions was 12.87\% for DenoFlow, compared with 34.75\% for Art. The same qualitative pattern was observed on Nakanishi2015, indicating that DenoFlow reduced the sensitivity of decoding performance to contamination severity within the evaluated range.
   \item DenoFlow did not completely restore the performance obtained with the uncontaminated reference trials. On Benchmark, the gap between DenoFlow and Ref. increased from 3.19\% under the mildest condition to 16.06\% under the most severe condition. Denoising reduces the cost of contamination but does not remove it.
\end{enumerate}

\subsection{Ablation Study}
Table~\ref{tab:ablation_results} presents the ablation study of the three loss terms. Each ablated variant removes one loss term while retaining the other two. The experiments were conducted on Benchmark at the reference condition, and the results were averaged across the three artifact types. The complete objective is included for comparison. Observe that:
\begin{enumerate}
    \item The complete objective achieved the best result for all four metrics, and each ablated variant performed significantly worse on every metric, with all Holm-corrected $p$ below $10^{-4}$ over the three variants. These results indicate that the three loss terms provide complementary supervision.
    \item The reconstruction loss made the largest contribution. Removing it reduced accuracy by 3.25\% and degraded all three fidelity metrics. This term provides direct sample-level supervision for the integrated output that is subsequently passed to the downstream decoders.
    \item Removing the classification and flow-matching losses reduced accuracy by 1.86\% and 1.70\%, respectively, and also resulted in smaller but statistically significant degradations in signal fidelity. These findings indicate that pointwise waveform reconstruction does not fully preserve task-relevant information or constrain the generation trajectory. The classification loss encourages the integrated output to retain discriminative information, whereas the flow-matching loss supervises the learned velocity field throughout the integration path rather than only at its endpoint.
\end{enumerate}

\begin{table}[!t]
	\centering
	\caption{Loss-term ablation on Benchmark, averaged over the three artifacts. Each of the first
	three rows drops one loss term and keeps the other two, and the last is the full objective.}
	\label{tab:ablation_results}
	\footnotesize
	\setlength{\tabcolsep}{2pt}
	\begin{tabular}{@{}ccc|ccc|c@{}}
		\toprule
		$\mathcal{L}_{\text{flow}}$ & $\mathcal{L}_{\text{rec}}$ & $\mathcal{L}_{\text{cls}}$ & CC & RRMSE & SNRI (dB) & Acc.\ (\%) \\
		\midrule
		\ding{55} & \ding{51} & \ding{51} & 0.8931$\pm$0.0660 & 0.4288$\pm$0.2135 & 7.70$\pm$1.20 & 72.99$\pm$22.21 \\
		\ding{51} & \ding{55} & \ding{51} & 0.8828$\pm$0.0603 & 0.4437$\pm$0.1941 & 7.24$\pm$1.28 & 71.44$\pm$22.67 \\
		\ding{51} & \ding{51} & \ding{55} & 0.8955$\pm$0.0596 & 0.4161$\pm$0.1821 & 7.77$\pm$1.26 & 72.83$\pm$22.18 \\
		\ding{51} & \ding{51} & \ding{51} & \textbf{0.9058$\pm$0.0543} & \textbf{0.3852$\pm$0.1498} & \textbf{8.23$\pm$1.36} & \textbf{74.69$\pm$21.06} \\
		\bottomrule
	\end{tabular}
\end{table}

\subsection{Parameter Sensitivity Analysis}
\begin{figure}[!t]
	\centering
	\includegraphics[width=0.8\linewidth]{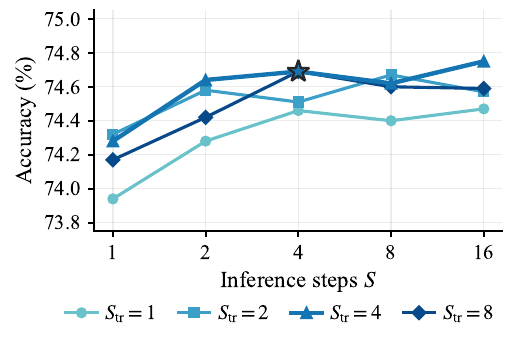}
	\caption{Sensitivity to the integration step counts on Benchmark, averaged over the three
	artifacts. Only $S=1$ is visibly worse. From two steps on, the curves sit on a plateau that
	contains the default (star).}
	\label{fig:sens_steps}
\end{figure}

Table~\ref{tab:sens_loss} examines sensitivity to the loss weights, whereas Figure~\ref{fig:sens_steps} evaluates different combinations of training and inference integration steps on Benchmark. All other hyperparameters were fixed at the default values specified in Section~\ref{sec:params}. The signal-fidelity metrics were averaged across the three artifact types, while decoding accuracy was additionally averaged across the five decoders. Observe that:
\begin{enumerate}
    \item Varying one loss weight at a time over $\{0.3,1,3\}$ yielded only modest changes in performance. Across all evaluated settings, the maximum accuracy difference was 0.67\%, while the ranges of CC, RRMSE, and SNRI were $0.0053$, $0.0150$, and $0.26$~dB, respectively. These results indicate that performance was relatively insensitive to moderate changes in any individual loss weight within the evaluated range, and no alternative setting consistently improved both fidelity and decoding accuracy.
    \item Across the complete grid of training and inference step counts in Figure~\ref{fig:sens_steps}, average accuracy ranged only from 73.94\% to 74.75\%. Changes in the training step count $S_{\mathrm{tr}}$ had little observable effect within the evaluated range. In contrast, using a single inference step ($S=1$) consistently produced lower accuracy across the tested values of $S_{\mathrm{tr}}$, whereas settings with $S\geq2$ formed a stable plateau. This pattern is consistent with diminishing numerical benefits from additional integration steps. The default configuration ($S_{\mathrm{tr}}=4$, $S=4$) lies within this stable region.
\end{enumerate}

Overall, DenoFlow remained stable under the evaluated loss-weight perturbations and integration-step configurations, indicating that its performance did not depend on a narrowly tuned hyperparameter setting.

\begin{table}[!t]
	\centering
	\caption{Sensitivity to individual loss weights on Benchmark. One weight was varied at a time while the other was fixed at $1$. Signal-fidelity metrics were averaged across the three artifact types, and decoding accuracy was additionally averaged across the five decoders. The $\lambda=1$ rows in the three blocks correspond to the same default run.}
	\label{tab:sens_loss}
	\footnotesize
	\setlength{\tabcolsep}{2pt}
	\begin{tabular}{@{}ll|cccc@{}}
		\toprule
		Parameter & Value & CC & RRMSE & SNRI (dB) & Acc.\ (\%) \\
		\midrule
		\multirow{3}{*}{$\lambda_{\text{flow}}$}
		 & $0.3$ & 0.9058$\pm$0.0516 & 0.3847$\pm$0.1410 & 8.19$\pm$1.37 & 75.08$\pm$20.92 \\
		 & $1$ & 0.9058$\pm$0.0543 & 0.3852$\pm$0.1498 & 8.23$\pm$1.36 & 74.69$\pm$21.06 \\
		 & $3$ & 0.9083$\pm$0.0537 & 0.3791$\pm$0.1430 & 8.33$\pm$1.44 & 74.41$\pm$21.13 \\
		\midrule
		\multirow{3}{*}{$\lambda_{\text{rec}}$}
		 & $0.3$ & 0.9109$\pm$0.0514 & 0.3702$\pm$0.1290 & 8.45$\pm$1.47 & 74.58$\pm$21.37 \\
		 & $1$ & 0.9058$\pm$0.0543 & 0.3852$\pm$0.1498 & 8.23$\pm$1.36 & 74.69$\pm$21.06 \\
		 & $3$ & 0.9056$\pm$0.0521 & 0.3844$\pm$0.1409 & 8.19$\pm$1.38 & 74.63$\pm$21.02 \\
		\bottomrule
	\end{tabular}
\end{table}

\subsection{Visualizations}
\begin{figure*}[htpb] \centering
	\subfigure[]{\includegraphics[width=0.32\linewidth,clip]{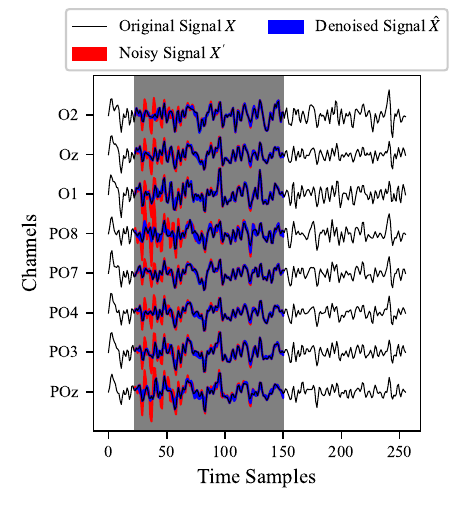}\label{fig:deno_emg}}
	\subfigure[]{\includegraphics[width=0.32\linewidth,clip]{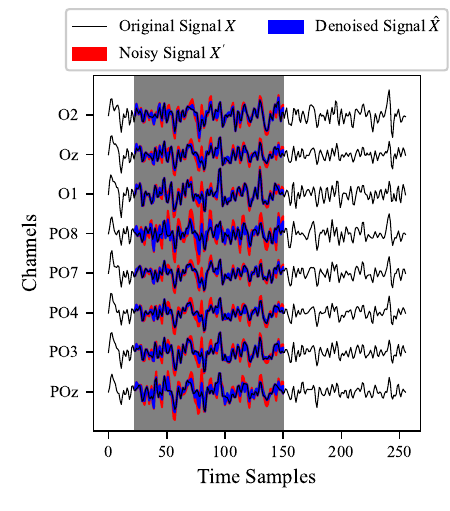}\label{fig:deno_eog}}
	\subfigure[]{\includegraphics[width=0.32\linewidth,clip]{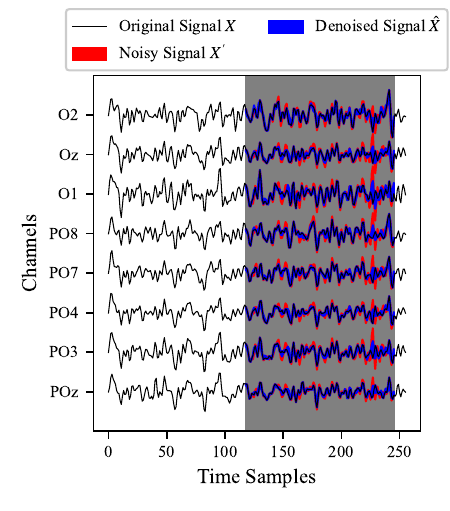}\label{fig:deno_both}}
	\caption{Qualitative denoising visualizations for the (a) EMG, (b) EOG, and (c) EMG-EOG mixed trials. The black, red, and blue traces denote the uncontaminated signal $X$, contaminated signal $X'$, and denoised signal $\hat{X}$, respectively. The shaded region indicates the artifact mask $M$.}
	\label{fig:deno_waveform}
\end{figure*}

Figure~\ref{fig:deno_waveform} shows qualitative denoising visualizations. The black, red, and blue traces denote the uncontaminated signal $X$, the contaminated signal $X'$, and the denoised signal $\hat{X}$, respectively, while the shaded region indicates the artifact mask $M$. Observe that:
\begin{enumerate}
    \item Within the masked intervals, the blue traces generally follow the black reference more closely than the red contaminated ones, indicating that DenoFlow substantially attenuated the injected artifacts while preserving the underlying waveform morphology.
    \item The reduction in the deviation from $X$ was most pronounced under EOG contamination and least pronounced under mixed contamination. This qualitative ordering is consistent with the correlation coefficients reported in Table~\ref{tab:fid_ds1}. As described in Section~\ref{sec:artifact}, the mixed condition combines ocular activity with a broadband muscular component and contains approximately $3$~dB more artifact power. Its greater spectral complexity and contamination level may therefore make signal recovery more challenging.
    \item The remaining discrepancies between $\hat{X}$ and $X$ were generally concentrated at time points where $X'$ deviated most strongly from $X$. This suggests that the reconstruction error depended on the local magnitude and temporal structure of the contamination rather than forming a uniform distortion throughout the masked interval.
\end{enumerate}

\section{Conclusion}\label{sect:conclusions}
This paper proposed DenoFlow, which casts SSVEP denoising as transport rather than as a direct mapping: A field network regresses the velocity of the straight path from a contaminated trial to its clean counterpart, and denoising integrates that field forward from the observation. The contaminated trial conditions the field network at every layer, and a jointly trained classifier supervises the integrated output. Training reduces to regression, without the adversarial min-max game of earlier generative denoisers. Experiments on two public SSVEP datasets with five popular SSVEP decoders showed that DenoFlow outperformed seven baseline denoising models on both signal fidelity and downstream decoding accuracy. Future work will explore extensions to online and cross-subject SSVEP decoding scenarios, as well as lightweight denoising model architectures for real-time deployment.

\section{Acknowledgments}
This research was supported by the National Natural Science Foundation of China under Grants (625B2077 and 62525305), and in part by the Open Fund of Intelligent Control Laboratory 2024-ZKSYS-KF02-04.

\bibliographystyle{model2-numbers}\bibliography{denoflow}

\end{document}